%% file: eccv12.tex
\documentclass[runningheads,a4paper]{llncs}

\usepackage{graphicx}
\usepackage{amsmath,amssymb} 
\usepackage{times}
\usepackage{epsfig}
\usepackage{color}
\usepackage{slashbox}
\usepackage{subfigure}
\usepackage[width=122mm,left=12mm,paperwidth=146mm,height=193mm,top=12mm,paperheight=217mm]{geometry}

\title{Abnormal Object Recognition: \\ A Comprehensive Study} 
\begin{document}
\mainmatter
\def\ECCV12SubNumber{1506}  

\author{Babak Saleh$^{1}$ \and Ali Farhadi$^{2}$ \and Ahmed Elgammal$^{1}$}
\institute{Rutgers University \and University of Washington}

\maketitle

\begin{abstract}
When describing images, humans tend not to talk about the obvious, but rather mention what they find interesting. We argue that abnormalities and deviations from typicalities are among the most important components  that form what is worth mentioning.  In this paper we introduce the abnormality detection as a recognition problem and show how to model typicalities and, consequently, meaningful deviations from prototypical properties of categories. Our model can recognize abnormalities and report the main reasons of any recognized abnormality. We introduce the abnormality detection dataset and show interesting results on how to reason about abnormalities.

\end{abstract}


\section{Introduction}
\input{Intro.tex}

\section{Abnormality: Problem Definition and Challenges}
\input{ProblemStatement}

\section{Related Work}

\input{relatedwork}

\section{Abnormality Dataset and Human Subject Experiments}
\input{Dataset.tex}

\section{Abnormality Detection Framework}
\input{classification}

\section{Experiments and Results}
\input{newresults.tex}

\section{Conclusions}
\input{conclusions}

\bibliographystyle{plain}
\bibliography{abnormality,psychologybib}
\end{document}

%% file: Intro.tex

The variability between  members of a category influences infants' category learning. 10-months-old infants can form a category structure and distinguish between category prototypes and atypical examples \cite{MatherPlunke2011}. 14-months-olds use properties of objects to report deviations from prototypes \cite{Horst2009}. In computer vision, there has been significant progress in forming the category structures. However, little attention has been paid to deviations from prototypical examples of  categories. This paper is centered on modeling the deviations from categories to be able to reason about atypical examples of categories and to detect abnormalities in images. Inspired by infant category learning, we propose to learn the structure of a category and then detect abnormalities as special deviations from prototypical examples.

There has been recent interest in investigating what should be reported as an output of a recognition system \cite{sentence2010}. When describing an image, human tend not to mention the obvious (simple category memberships) but to report what is worth mentioning about an image. We argue that abnormalities are among major components that form what is worth mentioning. We have probably heard statements like ``look at that furry dog,'' ``this is a green banana,'' several times. This type of reasoning is exactly what this paper is about. We want to form category structures in terms of common attributes in the category and reason about deviations from categories in terms of related attributes. Our method acknowledges  category memberships for atypical examples  and reports its reasoning behind any abnormality detection.

A diverse set of reasons may cause abnormality. An object can be abnormal due to the absence of typical attributes or the presence of atypical attributes. For example, a car with a plane wing is considered abnormal because typical cars don't have wings; a car without a wheel is also an atypical example of the car category because cars have wheel. Also,an abnormality can be caused by  deviations from the extent by which an attribute varies inside a category. Some examples include  a furry dog or a green banana. Furthermore, contextual irregularities and semantical peculiarities can also cause abnormalities; an elephant in the room~\cite{torralba2003iccv,torralba2005}.  In this paper we mainly focus on abnormalities stemming from the object itself, not from the context where the object is.  In Sec~\ref{S:Definition} we investigate the different sources of abnormalities and provide a definition of abnormality that we later use to develop our classification methodology. 

\noindent {\bf Why we need to study abnormalities in computer vision:} What studying abnormality in images tells us about object recognition? Humans seem to be able to categorize atypical instances of a given object class without learning on any of these atypical instances for humans. Categorizing an atypical instance of a class usually takes more time than typical instance~\cite{Eysenck05}. Can the state of the art computer vision object categorization and detection algorithms generalize as well to atypical images? 
We argue that testing categorization or detection algorithms on atypical images, without optimizing on them, provides insights on how these algorithms might simulate human performance. 


We aim to develop an  intelligent system that can detect and understand abnormality and should be able to perform the following tasks:
1) Detecting abnormality: detecting whether an image is abnormal  2) Categorization: recognizing the class of object despite abnormality
3) Identifying the reason for abnormality: abnormal shape, abnormal material, etc. 4) Quantifying the abnormality: defining a metric for abnormality is essential for many applications.

There are various applications for developing an intelligent system that can detect abnormalities. Certain types of abnormality in images can be an indication of abnormal event. This in particular is the case where an object in atypical pose with respect to the scene or within a unusual scene context. For example abnormal object poses and contexts in Fig.~\ref{F:examples}-Bottom are indication of abnormal event 


\noindent{\bf Contributions:}
There are different contributions for this paper: 1) The first in-depth study of abnormality in images covering different aspects of abnormality.
2) We propose a taxonomy of the causes abnormality. 3) The notion of abnormality is subjective. This makes the evaluation of abnormality detection more challenging.  We introduce an abnormality dataset for quantitative evaluation.  4) We introduce results of Human subject experiments that was designed to collect data about how humans decide about abnormalities. Such collected data is used as ground truth for our investigation.
 5) We introduce a framework based on combining discriminative and generative models for detection abnormalities in images.
 Our approach is based on learning the structure of a category and then detecting abnormalities as especial deviations from prototypical examples. 6) We investigate various state of the art techniques to evaluate their generalization performance on abnormal images and their ability to detect abnormalities

%% file: ProblemStatement.tex
\label{S:Definition}

\begin{figure}
\includegraphics[width=120mm,height=65mm]{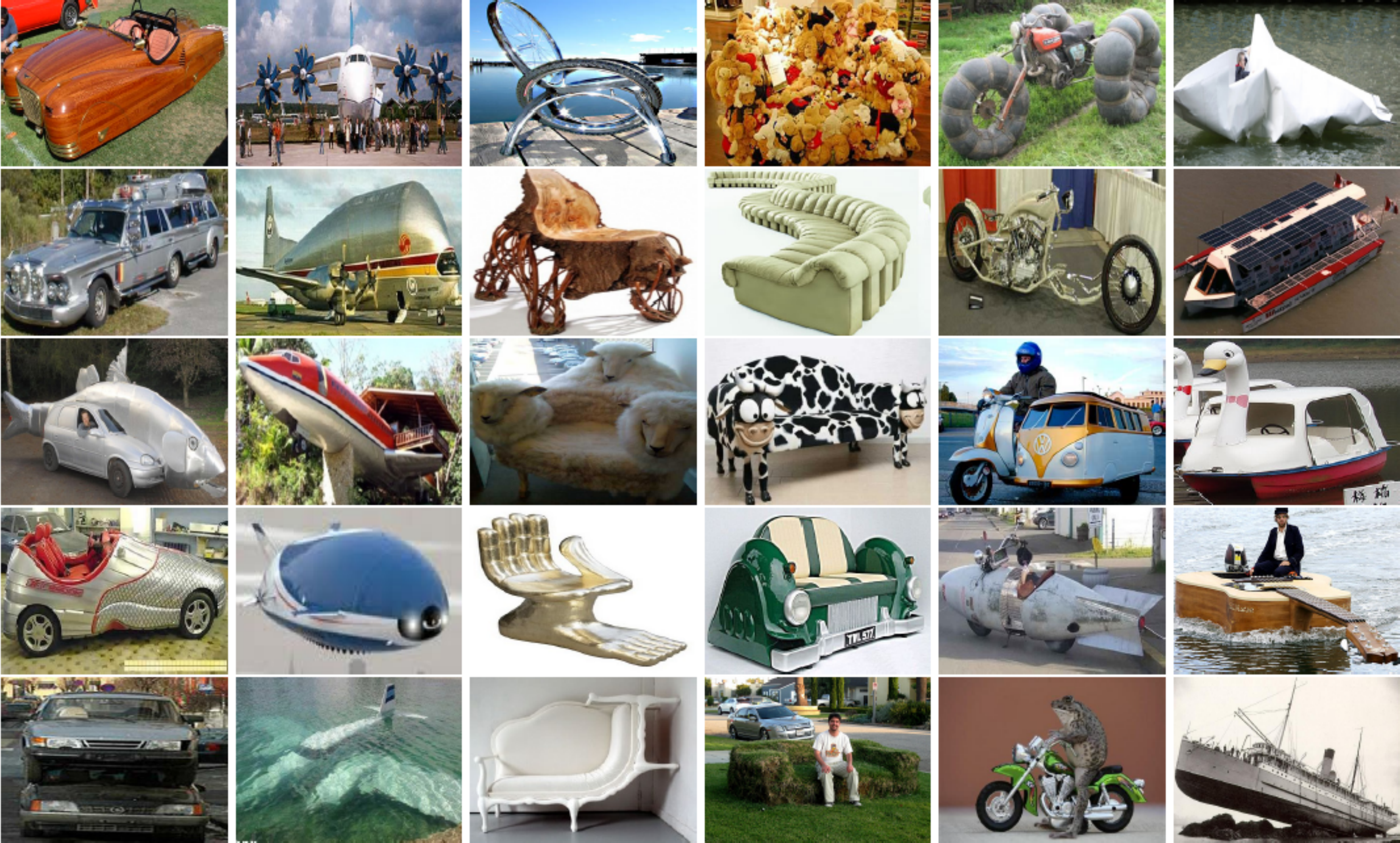}
\includegraphics[width=120mm,height=15mm]{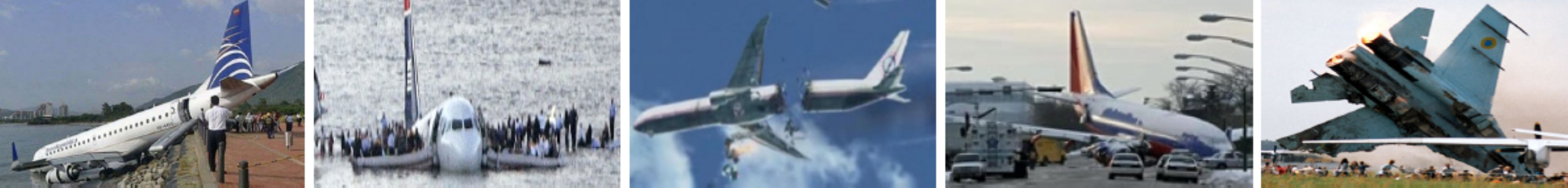}
\caption{Top Panel: Examples of abnormal images. Columns: images from six categories (cars, airplanes, chairs, sofas, motorbikes, boats). Rows: examples of different abnormality sources: atypical attributes, atypical shape, object is a combination of different categories, object is in the shape of another category, object is in atypical context or pose. Bottom Panel: Examples of abnormal object pose or context which is an indication of interesting events}
\label{F:examples}
\end{figure}
What makes an image to seem abnormal is a hard question. Definition of normality/abnormality depends on the subject's culture and personal experience. An image can be abnormal because it contains atypical instance of an object category. In such case humans might categorize the image correctly and might decide that it is abnormal. On the other hand, abnormality can also stem from confusion about the object category itself or the object relation to scene context. 

We can group the reason for abnormality to five groups; here we provide a  taxonomy based on our initial study of this subject. Providing such a taxonomy is essential for the developing of computational methods for detecting abnormalities. Abnormality can rise because of combination of these groups. The lines between these groups are sometimes blurred. Figure~\ref{F:examples} shows examples of six object categories from our exploratory dataset.
\begin{description} 
\item{I.} Abnormal attributes: For example atypical object texture, surface material, atypical parts, objects made of parts from other objects (e.g. first row of Figure~\ref{F:examples}.)
\item{II.} Abnormal object shape: Object shape or part configuration is atypical (e.g. second row of Figure~\ref{F:examples}.)
\item{III.} Object is a combination of multiple categories: here we can easily identify parts of the object as being from one category and other parts from another category, e.g., half a car and half a fish or half a car and half a boat. In most cases there is a dominant category, which we can identify based on the scene context, and atypical parts from a secondary category (e.g. third row of Figure~\ref{F:examples}.)
\item{IV.} Object in the shape of another identifiable object: For example, a car in the shape of shoe or a chair in the shape of a hand as in Figure~\ref{F:examples}. Here few parts and context information give us strong cues about the object category (functional category) while the configuration of the parts or the overall shape of the object indicates another secondary category (appearance category) - (e.g. fourth row of Figure~\ref{F:examples}.)
\item {V.} Abnormal object pose with respect to the scene or an object is in atypical scene context (e.g. fifth row of Figure~\ref{F:examples}. )
\end{description}

The distribution of object classes among the abnormality categories is not uniform. For example we can find many cars, chairs or sofas made in the shape of other objects or from atypical material; on the other hand we cannot find many airplanes that have such abnormalities.  This is expected and explainable since cars, chairs, motorbikes can be modified by humans much easier than airplanes. 

\medskip
\noindent{\bf Defining abnormality:} In psychology and cognitive science there is a long history of studying normality under categorization and prototyping theory (We review some related work in the next section), however what makes an image abnormal is not well defined.  
It is not easy to define abnormality. Let us consider the case where abnormality stems from the object itself and not because of context. Looking at the cars in Figure~\ref{F:examples} an abnormal instance of a car class should share enough resemblance (in terms of shape, appearance, or functional cues) to the car class such that we can tell it is a car, yet should be different enough that it triggers us to label it as abnormal. Therefore we define abnormal object as an object that is ''somehow'' similar to the class, yet has enough dissimilarity. Such definition might seem self contradicting since it talks about similarity and dissimilarity. If similarity is measured across one dimension, it would be very hard to find a threshold on similarity such that below it the object  is abnormal and above it the object is normal. This suggests that to judge about abnormality similarity has to be measured across different dimensions independently  and some consistency measures have to be employed.  

%

\medskip
\noindent{\bf Challenges:} How to approach abnormality detection: should we recognize object categories first and then detect abnormality? or should we recognize that the image is abnormal first and then categorize the object and/or the scene elements? 
Abnormality in many cases is identified in the context of an object category, e.g., abnormal attributes, or atypical shape. Defining abnormality as deviation from class norm dictates that categorization should be done first, prior to detecting abnormality. However in other cases an object might seem abnormal because it is a combination of different categories; or its category cannot be obviously categorized. 

In terms of classification, where does an abnormal instance of an object class exists in the feature space?  We can hypothesize that abnormal instances are on the margin between classes (e.g., the half-car half-fish example in Figure~\ref{F:examples}). But what makes such instances different from other typical instances close to the margin as well? So it is not clear how margin-based classifiers would be useful for categorizing abnormal instance, and how can they be used for detecting abnormality? It is also not clear how to model typicality in a way that enables atypicality detection, and how that  can be done in a multi-class setting?  It is clear though that we need multiple similarity measures, not just a simple distance-based classification approach. All these open issues makes the problem very interesting and challenging.

%% file: relatedwork.tex
\medskip
\noindent{\em Human judgments of typicality:}
The idea that members of categories vary in the degree of
typicality is fundamental to the modern psychological literature
on categorization \cite{Rosch_Simpson}, which is based on the
idea of family-resemblance structure in which category members
vary in their degree of fit \cite{Rosch_Mervis_75}. The exact
mechanism by which human learners determine typicality, or
determine category membership as a function of typicality within
a given category, is the main focus of most prominent theories of
human categorization. Some leading theories are based on exemplar
matching, similar to K nearest neighbor techniques
(e.g. \cite{Nosofsky_exemplar}) while others are based on central
prototypes and thus more akin to mean-of-class techniques
(e.g. \cite{Barsalou_1985}).  More recently the notion of
typicality has been put into a Bayesian framework
(e.g. \cite{Feldman_JEPHPP,Wilder_Feldman_Singh,Tenenbaum_typicality},
with typicality understood as the likelihood of the object
conditioned on the category. Nevertheless, the computational
mechanisms by which human observers assess visual typicality of
objects drawn directly from images remain an unsolved
problem.

\medskip
\noindent{\em Abnormality Detection:} The problem of abnormality detection for single images is not really well explored. Boiman and Irani~\cite{Boiman_detectingirregularities} studied the problem of irregularities in images and videos. In their definition, irregularities happen when a visual data cannot be composed by a large number of previously known data. They have shown interesting examples and applications in human activity recognition and detecting salient parts of images. 
Abnormalities and unusual behaviors have been studied in the context of human activity and event recognition in videos, e.g.~\cite{Ivanov99recognitionof}; however this is different from our goal.
Very recently, out-of-context objects have been studied in \cite{Choi2012} where a latent support graph has been learned to model the context.  Contextual irregularities is one of the reasons of abnormality in our taxonomy of abnormalities. Our goal is different, in this paper we mainly focused on abnormalities stemming from the object itself regardless of the context. In that sense our work is complementary to~\cite{Choi2012}.

\medskip
\noindent{\em Visual Attributes:} 
The choice of features upon which to determine typicality is
context-sensitive and depends on what features are considered
\cite{Schyns_Goldstone}.
The notion of attributes comes from the literature on concepts and categories (reviewed in~\cite{murphy2002}).  
The fluid nature of object categorization makes attribute learning essential.  For this reason, we make attribute learning the basis of our framework, allowing us to reason about abnormalities. Farhadi et al.~\cite{attr2009} and Lampert et al.~\cite{lampart2009} show that supervised attributes can be transferred across object categories, allowing description and naming of objects from categories not seen during training.  These attributes were learned and inferred at the image level, without localization. 
Attributes have been used as intermediate representation for object~\cite{attr2009,lampart2009}, face~\cite{facever_iccv2009}, and activity recognition\cite{liu-cvpr11b-attributes}. Recently, relative attributes have shown to produce promising results in recognition~\cite{relativeattr}. 
In this paper we adopt the attribute based representation of \cite{attr2009,attr2010}. We use global attributes in its standard form. However, as we are not focused on generalization properties of attributes, attributes are defined per category. We introduce an entropy based measure of attribute relevance for detecting abnormality within each category. This enables us to reason about abnormalities inside categories.


%% file: Dataset.tex


\subsection{Abnormality Dataset}
\label{Dataset}
For the purpose of our study, we needed to collect an exploratory dataset of abnormal images. We believe no such dataset exists in the computer vision community. There are datasets for studying abnormal activities in videos, however our goal is to study abnormalities in images.  To be inline with the image categorization research we chose object classes from PASCAL dataset~\cite{Pascal} to build our dataset. 
To collect the abnormal images in our dataset, we used image search engines, in particular Google images and Yahoo images where we searched for keywords like  ``Abnormal",  ``Strange",  ``Weird'' and ``Unusual'' in combination with class labels like cars, airplanes, etc. The top results from the search engines were pruned by removing duplicates, obviously irrelevant images and very low quality pictures.
Unlike typical images, it is not that easy to find abundance of abnormal images. Moreover abnormal images in some classes have obvious differences to what is in typical image datasets like PASCAL. For example, searching for ``abnormal people'' usually results in images with abnormal faces.  As a result we narrowed down the object classes to only six classes of PASCAL where we could collect at least 100 images: namely  ``Airplane", ``Boat", ``Car",``Chair",``Motorbike" and ``Sofa". The overall data set contains 617 images. The collected images were annotated by marking a bounding box around the salient object in the each image.

\subsection{Human Subject experiments}
The subject of abnormality is rooted in people's opinion, so any work on detecting strange images without any comparison to the human decision is not informative. There are other multiple reasons that motivates studying human subjects' responses to our collected images.  
1) Validating our collected dataset. 2) Providing ground truth 
3) Providing some insight about how people judge about the abnormality of images.  

Therefore, we designed a preliminary survey for human subjects and we used Amazon Mechanical Turk~\footnote{https://www.mturk.com/mturk/} to collect people responses. This might not be the most accurate and controlled subject experiment, however for the purpose of this paper we believe that this is sufficient. Given an image with a bounding box about the most salient object,  subjects were asked several questions. 
We collected 10 responses for each image.  
First the subjects were asked whether the image seems normal or abnormal. If the subject decides that the image is abnormal the following questions were asked where multiple selections are allowed:
1) Which category best describes the object, from a list of the six categories in our database. 
2) Whether abnormality is because of the object itself or its relation to the scene. 
3) To rate the importance of each of these following attributes in affecting their decision that this object is abnormal (Color,  Texture/Material, Shape/Part configuration,  Object pose/viewing direction) 4) Also the subjects were asked to comment about context abnormality if it is the case.
Typically the response time a subject takes to categorize an object is proportional to atypicality. However we could not measure these response times in this preliminary human subject study.

Fig.~\ref{F:Classification} shows the subjects' average rating for the different causes of abnormality for each category. This is for the images that subjects decide that the abnormality stems from the object itself. The figure clearly shows that in all categories atypical shape is the most common cause of abnormality, followed by texture/material, then pose and color. Except for the airplane category, the variances in the ratings  for each cause of abnormality is relatively small. The rating for the airplane has a large variance which might indicate that the real reason for abnormality is not one of the four reasons given. Fig.~\ref{F:Classification}-right shows the confusion matrix for the human subjects in deciding the categories.  An important conclusion from this study is that the variance in subjects' decisions about Normality/Abnormality is much less than the variance in their decisions about the object categories.

%% file: classification.tex
\subsection{Abnormality Classification Paradigm}

\begin{figure}[t]
\parbox{5.1in}{
\parbox{2in}{
\includegraphics[width=2in,height=1.5in]{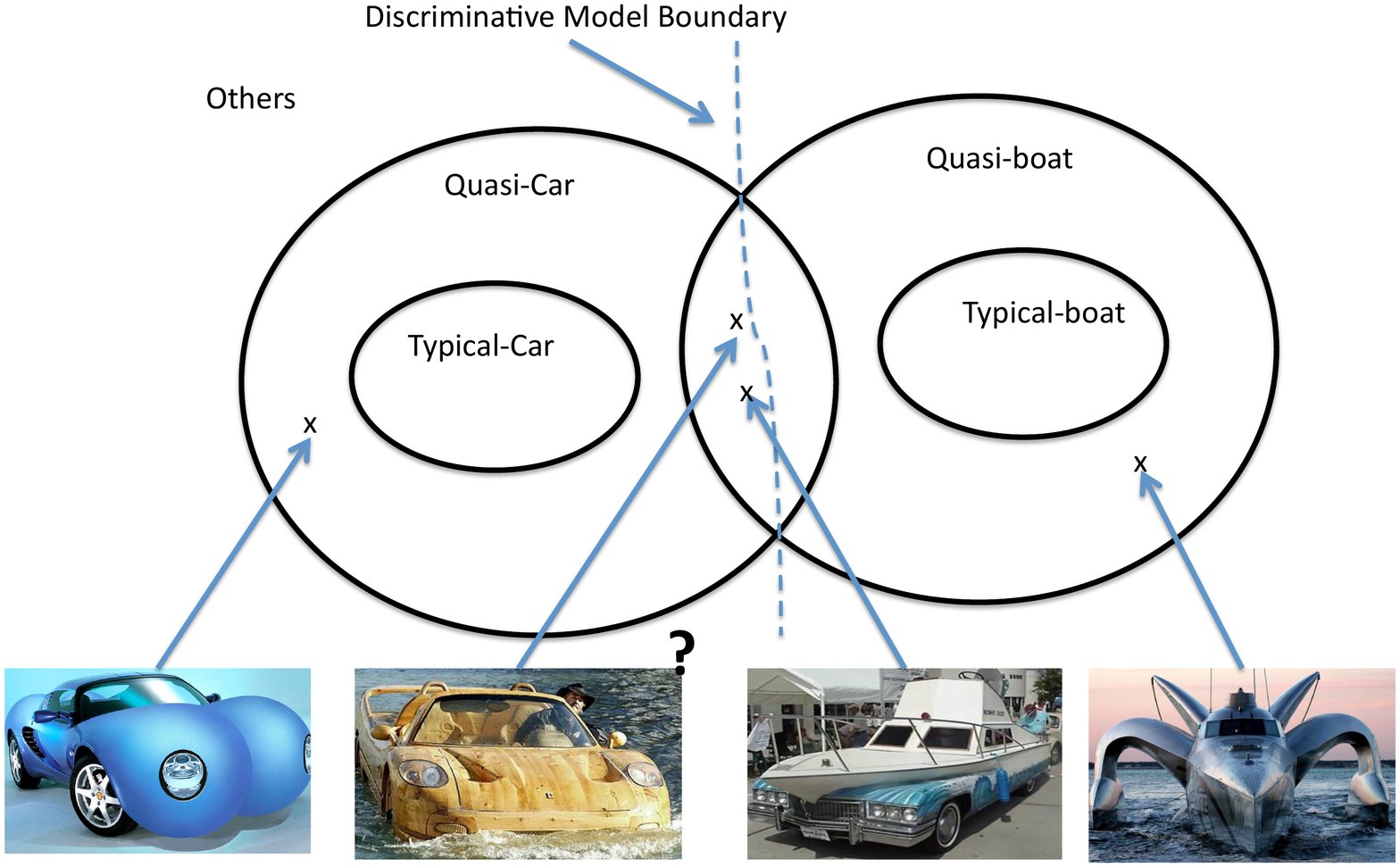}
}
\parbox{1in}{
\includegraphics[width=1in,height=1.5in]{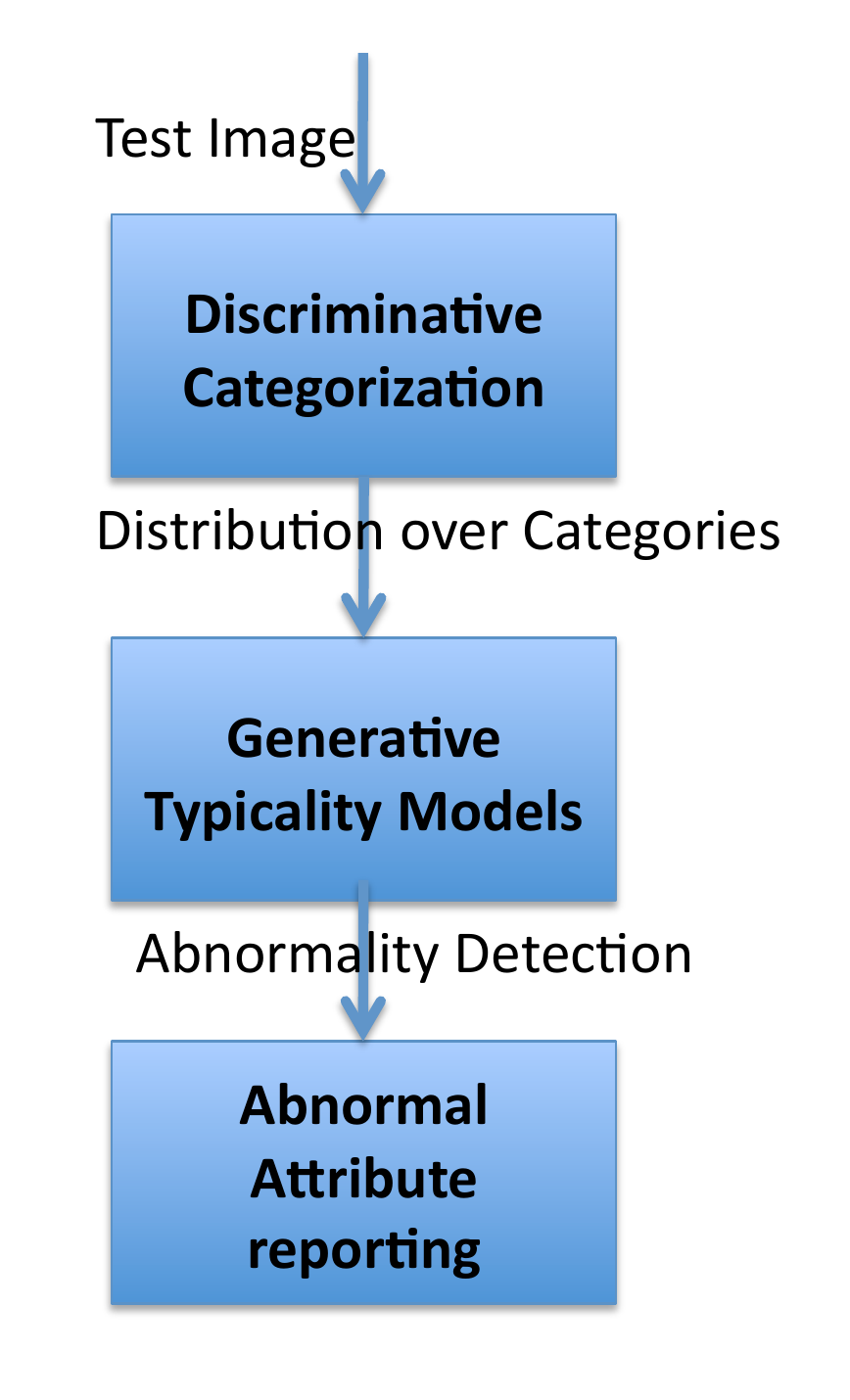}
}
\parbox{2in}{
\includegraphics[width=1.5in,height=0.6in]{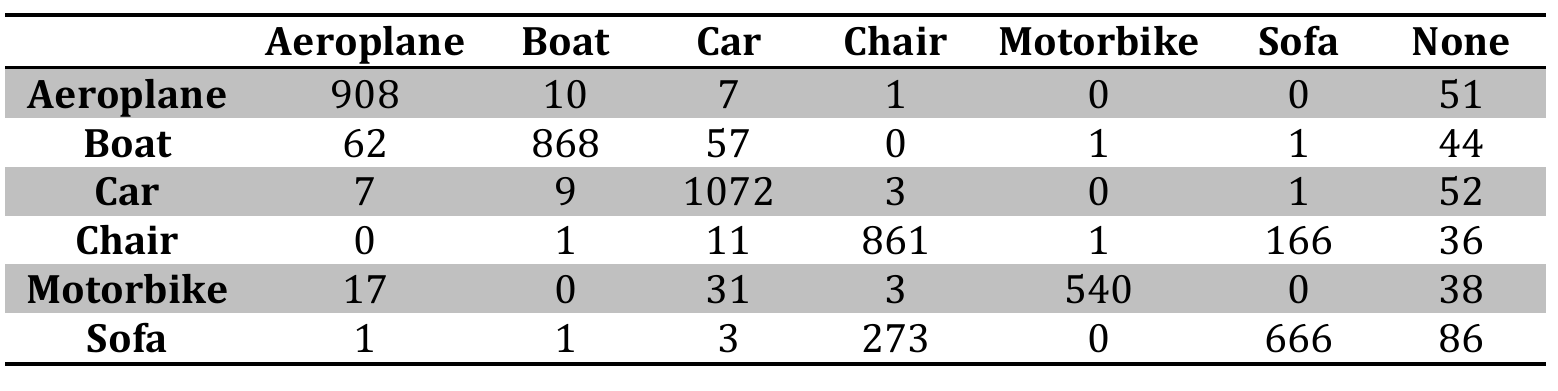}
\includegraphics[width=1.8in,height=0.9in]{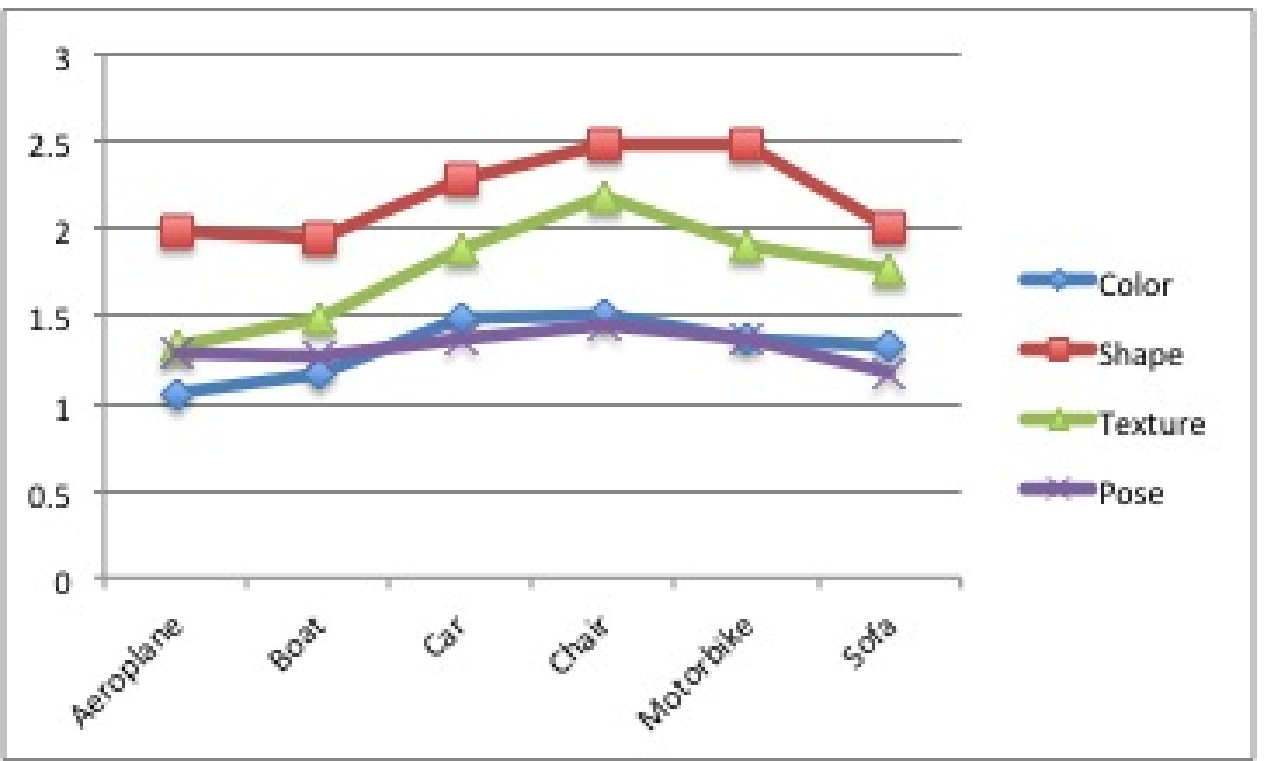}
}
}
\caption{Left: Illustration of car/boat classification with abnormal images. Right-top: Categorization Confusion Matrix for Human Subjects. Right-bottom: Subject's rating of different sources of abnormality}
\label{F:Classification}
\end{figure}

There are two observation that motivates our classification model. First, the common approach to multi-class recognition involves performing several one-versus-all classification/detection tasks. Such a discriminative paradigm shows superiority in categorization and thus widely used. This implies that there is an assumption about existence of  clear boundary between object categories. Taking abnormalities into consideration,  the boundaries between basic level categories become not as clear. In particular, objects in group III,IV in the abnormality taxonomy, contains several features and attributes that are common to multiple classes. It might be hard or impossible to identify the correct category of these objects mainly based on visual features or attributes of these objects themselves. Further investigations of the scene context and the functionality of these objects are necessary to determine the category of these objects. Therefore the outcome of the categorization phase in a recognition system should not be a hard decision of category, rather membership scores of different categories. 

Second There is a fundamental difference between our definition of abnormalities and the existing definitions in the literature. In conventional definitions unusual examples are the ones that are not similar to any (or similar to very few) previously known examples. However, our definition of abnormality (as discussed in Sec~\ref{S:Definition}) entails a form of similarity while being different. For example, based on conventional definitions, a chair can be thought of as an atypical example of car category, atypical example of motorbike category, etc. In contrast, our definition of abnormality requires the example to be ``some how'' similar to some categories while being different in some related attributes. 

Where do abnormal instances of categories lie in a visual feature space? The above two observations lead to the following hypothesis. For each category we define two different sets: the set of normal/typical instances, and the set of quasi-category. The set of quasi-category contains the instances that resembles the category in certain features or attributes however they are atypical from category prototypical examples. This is illustrated in Fig~\ref{F:Classification} where we use two categories for simplicity, car and boat. There are the sets of typical-cars and typical-boats which are disjoint; and there are also the sets of quasi-cars and quasi-boats. The typical-category set is a subset of the quasi-category set. The quasi-category sets can intersect and do intersect in many cases, e.g. there are instances that resembles cars and boats that belong to the intersection of the quasi-car and quasi-boat sets. 

The typical discriminative categorization algorithms do not consider this setup, and assume a clear boundary between categories. So they are bound to be confused about instances in the quasi-category intersections. Human also get confused about these instances as apparent form the confusion matrix in Fig~\ref{F:Classification}. This might suggest detecting abnormal instances based on their closeness to the margin. However, this is not sufficient since abnormal instances can also be away from the margin, anywhere in the quasi-category set; see the baseline experiment in Table~\ref{T:Typicality}.

Therefore, to be able to detect abnormal instance of category $c$ we need some indicator that this instance is in quasi-$c$ and not in typical-$c$. However, the challenge is that the boundary of typical-$c$ and the quasi-$c$ is not well defined, as well as the boundary between quasi-$c$ and the rest of the world. Furthermore we should not train on abnormal instances (humans do not train on abnormal images), therefore discriminative approach for detecting abnormality within class is neither feasible, nor desirable. 

The above discussion makes it clear that a generative model is needed to model typical instances of a given category. Therefore we propose a two stage approach where at the first stage a discriminative model is used for categorization. Our model produces a distribution over categories and avoids making hard decisions till the very end in the process.  At the second stage a generative model is used to detect typicality within each category. Deviation from typicality within the category should indicate abnormality. This process can be formally defined in a simple probabilistic form: given an instance $x \in \mathcal{X}$ where $\mathcal{X}$ is an arbitrary feature space, the first stage produces a distribution over categories, i.e., $p(C | x)$ where $C=\{c_1\cdots,c_K\}$ is a random variable indicating the category. At the the second stage a typicality model is learned for each category in terms of the a class-conditional density model  $p(x | T, c_k)$ from typical instance. Here we introduce another random variable $T$ to indicate typicality. 
Theoretically by Bayes' rule the posterior $p(T|x,c_k)$ could be achieved, and the atypicality given the category is simply the complement event. Thus we can obtain $p(T , c_k  |x) =  p(T | c_k,x) p(c_k |x) $. However, in our case we cannot get the posterior because we do not have a model for $P(x | \neg T,c_k)$, which is the generative model for the atypical instances. This is because we should not train on atypical instances. Therefore we have to use the likelihood to decide about typicality. 

Our proposed approach models the typicalities by leveraging the hidden structures among typical examples of categories using an attribute-based representation. Once the typicalities have been modeled, abnormalities can be defined as meaningful deviations from typicalities within the category. 
To model this deviation one needs to encode related attributes and select accordingly. Once deviations have been formulated, our method can classify atypical examples, and reason about the rational behind any detection in terms of attributes. To achieve this goal we need to 1) investigate generative methods for discovering the structure of typicality, 2) devise methods to measure deviations from typicality.

Unlike most attribute based frameworks, our attributes are not designed to provide cross category generalization. In fact, we intentionally learn our attributes to encode inside category relationship. Because, there are subtle differences for attributes inside categories. A typical bicycle wheel is considered atypical for cars. Furthermore, patterns of occurrence of attributes may be very different for very similar categories. Later we show how to benefit from these patterns of co-occurrences.

\subsection{Relevant Attribute Selection}
An attribute is useful for detecting normality/abnormality if it is common with a given category. For example, cars typically have wheels, if a car in an image does not have wheels and there is no obvious reason for not seeing the wheels, then it is probably abnormal. On the other hand an attribute is useful for detecting abnormality if it is rarely seen in a given category. Take the car example again, a car is not expected to have wings or eyes. Existence of such attributes are a strong cue of abnormality. So the absence of  common attributes or existence of  peculiar attributes for each category are useful cues for detecting abnormality. 

Let $A_i(x): \mathcal{X} \rightarrow \mathbf{R} $ be the confidence of the $i$-th attribute obtained from the $i^{th}$-attribute classifier for image $x$. We need to model the conditional density $p(A_i | T, c_j)$ for each attribute $i$ given typical example of category $j$. Both common attributes and peculiar attributes share the properties that they should have a peaky conditional densities, regardless of the value of the confidence. Therefore, we use an entropy measure to detect such attributes. We compute the conditional entropy $H(A_i | T, c_j)$ for each attribute and category pair. The lower the entropy the more peaky the distribution of the confidence over typical images, and hence the more relevant that attribute for detecting typicality/atypicality. Therefore we use $ 1/H(A_i | T, c_j)$ as the typicality/atypicality-relevance measure of attribute $i$ for category $j$.
 
\subsection{Modeling Typicality}
\label{S:Typicality}
For modeling typicality we need learn generative models in terms of the conditional class densities $p(x|T,c_k)$. We use an attribute space for that purpose, i.e. we need to model $p(A_1(x),\cdots, A_M(x) | T, c_k)$, where $M$ is the number of attributes. We investigated several models of typicality, which we will summarize in this section. 

\medskip
\noindent{\bf Naive Bayes' Model:} In this approach we model the density $p(A_1(x),\cdots, A_M(x) | T, c_k)$ = $\prod_i p(A_i(x) | T, c_k) $ where we use a Gaussian model for each attribute density :
\newline $ p(A_i(x) | T, c_k) $ $\sim \mathcal{N}(\mu_i^k , {\sigma_i^k}^2)$. 

\medskip
\noindent{\bf Nonparametric Model:} In this approach we model each conditional class density using kernel density estimation, i.e., we achieve an estimate of the density in  the form $\hat{p}(A_1(x),\cdots, A_M(x) | T, c_k) = \frac{1}{N} \sum_{j=1}^{N} \prod_{i=1}^M  g(A_i(x)-A_i(x_j))$, where $g(.)$ is a kernel function and $\{x_j\}$ are training images of class $k$. Here we use the kernel product, which is typically used to approximate multivarite densities.

\medskip
\noindent{\bf Modeling typicality manifold:} In this approach we hypothesize that typical images lie on a low-dimensional manifold in the attribute space. We explicitly model that typicality manifold for each category and compute deviation from abnormality by modeling the distance of a test image to that manifold. Given a test image we find its nearest neighbor from the training data of a given category and then compute the perpendicular distance to the tangent space of the manifold at that point. This can be achieved by projecting the test image to a local subspace for the manifold patch around the nearest neighbor point. There are two probability models for the distance to the manifold that we investigated:  1) a global Gaussian model for the whole manifold, i.e.,   2) a local Gaussian model at each patch of the manifold. There are two parameters for this model, the patch size, $k$ and the local subspace dimensionality $d$. 
 
\medskip
\noindent{\bf Manifold-based density model:} This approach is similar to the Naive Bayes' Model, however instead of computing the densities $ p(A_i(x) | T, c_k)$ globally, these densities are computed locally for patch of the typicality manifold. The rational is each part of the typicality manifold is expected to have different distribution. 
 
 \medskip
 \noindent{\bf  One-class SVM:} One-class SVM is typically used for estimating regions of high density. Given typical examples for each class in the attribute space, one-class svm is used to estimate a boundary of volume of high density, which is then can be used to detect deviation. 
 
 \subsection{Reporting Abnormal Attributes}  
 \label{S:AttReporting}
Each recognized abnormality can be supported by reasons of the abnormality in terms of attributes. We use an information-theoretic measure for detecting abnormal attributes. Measuring the information content of a given attribute is an indication of the rareness or commonality of that attribute. However, in order to compute the information content, we use the manifold-based density model as described above. The different attributes are weighted according to their typicality/atypicality-relevance scores.  Finally attribute information content are aggregated according to different subsets of attributes (shape, texture, color, object pose). Given the categorization decision $c^*$, the information content for a subset $S$ of attribute can be computed as
$  \sum_{i \in S}  \frac{1}{H(A_i | c^*)} \: \log_2 \frac{1}{P_{\mathcal{M}} (A_i(x) |T,c^*)} $
where $P_{\mathcal{M}} (A_i(x) |T, c^*)$ is the likelihood given the manifold-based local conditional density.

%% file: newresults.tex
Because of the lack of space we mainly describe the quantitative results in the paper, many qualitative results and more quantitative results are available in the supplementary material.

\subsection{Features and Attributes}
We describe and model objects using visual attributes, which can be categorized into shape, color, texture and part related attributes. To learn a broad range of attributes we need a wide variety of features, which  we call "base features". Similar to \cite{attr2009,attr2010} we use edges to model the shape, and pyramid of Histogram of Oriented Gradient features to find part attributes. ColorSIFT and Texture features are extracted to learn attributes which are related o material and texture. Base feature extraction has been done in a pyramid-based approach. we divide the image into six patches and extract base features for each of these patches in addition to the whole image. We apply canny edge detector, quantized output of HoG and Texton filter bank responses. Also, unlike\cite{attr2009,attr2010} we use ColorSIFT to improve features for learning attributes related to color and material. This feature extraction process will end in a 10751 dimensional feature vector for each image. 
We use 64 visual attributes, each one is learned using a SVM classifier on top of selected dimensions of base feature vector. To find out that dimensions of base feature vectors are important for a specific attributes, we fit a $l_1$-regularized logistic regression between objects coming from a specific class with that attribute and without it.

\subsection{Evaluation of State-of-the-art Methods}

To get an insight on how the state-of-the-art algorithms behave on abnormal images, and to obtain baselines, we performed several evaluation of state-of-the-art algorithms for the following  tasks. This evaluation is also fundamental to our approach since we use the categorization result as the first stage in our approach.

\noindent{\em - Detector:} 
We use the state of the art deformable part-based detectors of~\cite{pedro2010} to evaluate how well one can categories images of abnormal objects. Here, we don't care about localization based measures. Therefore we relax the overlap constrain to zero. This means that we want to use this detector as a classifier and it would be a correct response if the detector fires on an image that contains instances of desired category. We hypothesize that this approach should fail when applied on abnormal images because abnormal images do not exhibit normal part configuration. Numbers in Table~\ref{T:Evaluation} shows the percentage  of the cases where the detector could do classification correctly.

\noindent{\em - Categorization - Base features:} Each image is represented using base features and one-vs-all SVM classifier is trained for each category.

\noindent{\em - Categorization - Attribute based classifier~\cite{attr2009} for categorization:} Each image is represented by a feature vector which is the output of 64 attribute classifiers. We trained two different classifiers: one-class SVM classifier for each category and a one-vs-all  SVM classifier. The one-class SVM only trains on positive examples of each class. 
In all cases the models were trained on subsets of PASCAL images (denoted as the normal dataset) and no training is done on the abnormal dataset. For part-based detectors we used the trained models provided by the authors~\cite{pedro2010} (also trained on PASCAL). We evaluated on both the normal (600 images from PASCAL test) and our abnormal dataset. The results are shown in Table~\ref{T:Evaluation}. It is surprising to see the large divergence in the results, while part-based detectors failed, as expected to detect the objects, the attribute-based  and the base-features categorization approaches is consistently able to categorize the abnormal images. Of course the performance on categorizing abnormal images is not as good as the case of normal images in most of the cases, which is expected, but the generalization to the unseen abnormal images is quite surprising. There are even cases where the performance on the abnormal images is better than the normal test images. There are various conclusions and observations we can make out of this experiment. First we can reject the hypothesis that the bad performance for part-based detectors is because of different bias in the abnormality dataset, since the categorization approaches performed consistently on it. Second, failure of part-based detectors might be used as a strong cue of abnormality in an image given that we actually have another way to detect the object and correctly categorize it! Third, it is clear that the attribute-based approach captures a good representation of each category that carried over for unseen test instances from both the normal and abnormal datasets. 

\begin{table}[t]
\caption{Evaluation of different approaches for categorizing abnormal images. Percentage accuracy is shown.}
\label{T:Evaluation}
\centering
\begin{center}
\footnotesize
\resizebox {0.9\textwidth }{!}{
\begin{tabular}{|l|l|l|l|c|c|c|c|c|c|c|}
\hline
Task & Method & Features &test dataset & Airplane &Boat&Car&Chair&Motorbike&Sofa\\
\hline\hline
Categorization & one-vs-all SVM & base & PASCAL & 81.83 &	74.67 &	76.67& 	81.0 &	81.5	& 81.5 \\
\hline
Categorization & one-vs-all SVM & base & Abnormal & 55.92 &	75.69 &	68.23& 	72.77&	64.67 &	46.03 \\
\hline
Categorization & one-vs-all SVM & Attributes & PASCAL & 78.66 &	61.17& 	63.33 &	65.33& 	77	& 82.17 \\
\hline
Categorization & one-vs-all SVM & Attributes & Abnormal & 58.99 &	64.67 &	70.65	& 73.09	&  73.58	& 64.18 \\
\hline
Categorization &  one-class SVM  & Attributes &PASCAL &76.85&77.45&76.55&77.59&75.09&76.37\\
\hline
Categorization & one-class SVM & Attributes &Abnormal  &71.05&69.50&59.90&67.99&67.28&63.65\\
\hline
\hline \hline
Detection & Part-based & HoG & Abnormal  &5 \%&3 \%&35 \%&0 \%&10 \%&0 \%\\
\hline
\end{tabular}
}
\end{center}
\end{table}

\subsection{Evaluation of Normality/Abnormality classifiers}
We evaluated the various proposed methods for modeling typicality given the category as described in Sec~\ref{S:Typicality}. For all these experiments we trained the typicality models using the same training data from PASCAL train. The number of images per class varies as indicated in Table~\ref{T:Typicality}-I. For testing we used a mixture of normal images  from PASCAL (100 per class) and abnormal images from our dataset (100 per class). Since the goal is to evaluate the Normality/Abnormality classifiers given the class, out of these test images we only used the ones that are correctly categorized by the first stage categorization. The baseline for this experiment is a typicality model learned on the result of the first stage categorization classifier. We used the confidences from the one-vs-all svms used for categorization and fit a Gaussian model for the distribution of the confidences for the typical images of each class. We use this Gaussian Model to obtain a probability of being typical given the category.

The AUC results for the Normality/Abnormality classifiers for each category are shown in Table~\ref{T:Typicality}-I. On average the Naive Bayes approach with attribute relevance give the best results, with almost similar result using the one-class svm. 
The global manifold distance model gives the best results for the Car and Chair categories where there are a lot of training samples, while it does not perform as well for the categories with small number of samples. This is expected since any manifold approach needs a dense sampling of the underlying manifold. We hypothesize that the manifold model should give the best results if all categories have enough training data.

\begin{table}[t]
\caption{\footnotesize Normality/Abnormality Classification Results}
\label{T:Typicality}
\vspace{-0.5cm}
\centering
\begin{center}
\resizebox {0.80\textwidth }{!}{
\begin{tabular}{| l | c | c | c | c | c | c | c | }
\multicolumn{8}{l}{ \bf I. Normality/Abnormality Classification within each category (AUC)} \\
\hline
\backslashbox {Approach} {Object class}&Airplane&Boat&Car&Chair&Motorbike&Sofa & Average \\
  &  270 & 353 &  922 & 811 & 197 & 153 & \\
\hline\hline
Baseline  & 0.5183 &	0.7397 &	0.5671 &	0.9211 &	0.6682	& 0.6011 & 0.5597 \\
\hline
{Naive Bayes}& 0.6230& 0.9394& 0.8847& 0.9882& 0.8136& 0.7149 & 0.8273 \\
\hline
Naive Bayes with Attribute relevance & 0.6638 & {\bf 0.9403} & 0.9166 & 0.9876 & 0.8021 &0.6919 & {\bf 0.8337 }\\
\hline
{Nonparametric Model}& {\bf 0.7265} &0.7917&0.6629&0.5057& {\bf 0.8681} & {\bf 0.7963} & 0.7252 \\
\hline
Global Manifold Distance & 0.5280 & 0.8887 & {\bf 0.9318 } & {\bf 0.9901 } & 0.7437 & 0.6429 & 0.7875\\
\hline
Local Manifold Distance & 0.5771 & 0.7480 & 0.7376 & 0.9404 & 0.7437 & 0.7196 & 0.7444\\
\hline
Manifold-based Density Model & 0.6406 & 0.8196 & 0.7990 & 0.9218 & 0.7009 & 0.6238  &0.7510 \\
\hline
{One class SVM}&0.6615&0.9370&0.9222& {\bf 0.9901} &0.8140&0.6693 & 0.8324 \\
\hline
%
\multicolumn{8}{l}{\bf II. Evaluation of abnormal attribute reporting - KL divergence from ground truth} \\
\hline
\backslashbox {Approach} {Object class}&Airplane&Boat&Car&Chair&Motorbike&Sofa & Average\\
\hline\hline
Baseline(1) &0.0796&0.08&0.0775&0.1035&0.0944&0.064 &0.0832\\
\hline
Baseline(2)&0.0826&0.0768&0.0809&0.0956&0.0892&0.0565 &0.0803\\
\hline
Our Approach &0.05669&0.03689&0.07583&0.06315&0.06349&0.06954 &0.0609\\
\hline
\end{tabular}}
\end{center}
\end{table}

\subsection{Abnormal Attribute Reporting}
We used the collected human responses for abnormality rating and abnormality source rating as ground truth. The output of the abnormal attribute reporting  is compared to the ground truth using correlation analysis. We use Kullback-Leibler divergence to compare human's rating vs. the output of the algorithms. Alternatively we can compare ranking divergence. 
Our final contribution will be detecting abnormal aspects of the object and report them in terms of visual attributes. To be able to quantitatively compare our results to what people think of abnormal attitudes of an object in Amazon TURK experiment, we grouped 64 attributes into four exclusive categories: Shape, Material/Texture, Color and Object pose. 
In this experiment baselines are based on Farhadi et al\cite{attr2009}. First row of Table~\ref{T:Typicality}-II is regarding to experiment when we take the mean and variance of each attribute for all the training data from a specific class. So we get 64 dimensional attribute mean $\mu_{i}$ and variance $\sigma^{2}_{i}$ for each class $i$.
For each test image after categorization and abnormality detection, if the image is an abnormal instance of class $j$ then we will focus on detecting abnormal attributes. Attribute$i$ in test image will be abnormal for class $j$ if the attribute confidence does not fall in the range of $2*\sigma_{i}$ around the attribute mean for class$j$. Baseline 2 is similar to previous try, but this time we increased the range interval to $4*\sigma_{i}$
We evaluated the abnormality reporting using our approach as described in Sec~\ref{S:AttReporting}. 

%
%
%
%
%
%

%% file: conclusions.tex

In this paper we presented results of our investigation on the subject of abnormality in images. We introduced a dataset for abnormal images for quantitative evaluation along with human subjects' ground truth. We propose a taxonomy of the reason of abnormality which should be helpful in algorithmic development. We mainly focused on abnormality stemming from the object itself.  We proposed and test a classification paradigm that is suitable for the problem of abnormality detection. We showed that attribute-based categorization methods generalizes well to abnormal images without the need to train on them. We also showed a generative model of typicality within each category is needed to jude abnormality and detect the source of it. This study just scratch the surface of this topic, there are many open questions including: how can detecting abnormality be useful in categorizing confusing objects? How to combine different abnormality cues?